\documentclass[letterpaper, 10 pt, conference]{ieeeconf}  

\IEEEoverridecommandlockouts                              

\usepackage{amsmath}
\usepackage{subcaption}
\usepackage{graphicx}
\usepackage{booktabs}
\usepackage{graphics} 
\usepackage{epsfig} 
\usepackage{mathptmx} 
\usepackage{times} 
\usepackage{amsmath} 
\usepackage{amssymb}  
\usepackage{float}
\usepackage{algorithm}
\usepackage{algorithmicx}
\usepackage{algpseudocode}
\usepackage{subcaption}
\usepackage{xcolor}
\usepackage{url, hyperref}

\title{\LARGE \bf
A Surveillance Evasion Game with Continuous Sensor Redeployment via Bilevel Optimization
}

\author{Jaehyeok Kim$^{1}$, Kartik A. Pant$^{2}$, Joseph Kinerson$^{3}$, Kylie Sommer-Kohrt$^{4}$, \\ Worawis Sribunma$^{5}$, Li-Yu Lin $^{6}$and James M. Goppert $^{7}$
\thanks{This material is based upon work supported by the Office of Naval Research (ONR) under the Threat and Situational Understanding of Networked Online Machine Intelligence (TSUNOMI) program (grant no. N00014-23-C-1016). The authors also thank Dr. Christopher Vo and his team at SAAB Inc. for their feedback and support.}
\thanks{The authors are with the School of Aeronautics and Astronautics, Purdue University,
West Lafayette, IN 47906. Email: (kim2153, kpant, jkinerso, ksommerk, wsribunm, lin1191, jgoppert)@purdue.edu}
}

\begin{document}
\maketitle
\thispagestyle{empty}
\pagestyle{empty}
\begin{abstract}
Uncrewed Aerial Systems (UASs) have become a growing threat to the security of critical infrastructure, exploiting spatiotemporal gaps in sensor perimeters to infiltrate restricted airspace undetected. We formulate this interaction as a two-player zero-sum differential game between an adversarial UAS and a heterogeneous sensor network of directional and omnidirectional sensors. Unlike earlier game-theoretic approaches that restrict the defender to discrete placement graphs or fixed configurations, we introduce a continuous sensor redeployment technique in which each sensor slides freely along the convex building boundaries. This is enforced via a log-sum-exp smooth approximation that preserves differentiability at polygon vertices, enabling optimization with gradient-based methods. The attacker's best response is computed via a two-step approach combining STP-RRT* for feasible trajectory initialization and nonlinear programming for detection-minimization refinement.
The joint optimization converges to a Local Nash Equilibrium (LNE) via alternating bilevel optimization, with analytical first-order stationarity conditions derived for both players, thereby establishing a deployable baseline for heterogeneous sensor placements in CUAS missions.
A 500-trial Monte Carlo simulation demonstrates that the proposed framework achieves a $4\times$ improvement in defender detection probability over randomized sensor placement, the natural baseline given the absence of directly comparable prior methods, with a $96.8\%$ convergence rate, confirming convergence toward a competitive LNE.
\end{abstract}

\section{INTRODUCTION}

Increasing incident reports of Uncrewed Aerial Systems (UASs) violating restricted airspace \cite{UAV_airspaceviolation} and being used for covert surveillance \cite{UAV_cyberattack, UAV_privacy} highlight a growing threat to critical infrastructure. Commercially available platforms for recreation or delivery are modified for adversarial misuse. Airports, military installations, and public venues face safety threats, as a single undetected intrusion can have severe consequences. Addressing this threat requires a principled framework that models the adversarial interaction between an intelligent intruder and an adaptive sensor network, enabling optimal defensive deployment under realistic operational constraints.

In a typical Counter-UAS (CUAS) mission, a defender secures a perimeter with assets subject to legislative and budgetary restrictions, e.g., an airport security team limited to a fixed sensor budget and prohibited from deploying equipment beyond designated infrastructure boundaries. Strategic deployment of defender assets is traditionally viewed as optimal sensor placement \cite{springer_optimal_camera_placement, thesis_optimal_camera, CUAS_03}. However, present approaches commonly assume sensors are either static or omnidirectional, detecting threats uniformly in all directions. In practice, high-gain directional sensors rotate periodically, producing a time-varying and spatially limited coverage region \cite{fov_ref}. This creates a predictable spatiotemporal gap, a window in both position and time when no sensor is surveying, that an intelligent adversary could exploit with precise timing and path planning.

\begin{figure}[t!]
    \centering
    \includegraphics[width=0.95\linewidth]{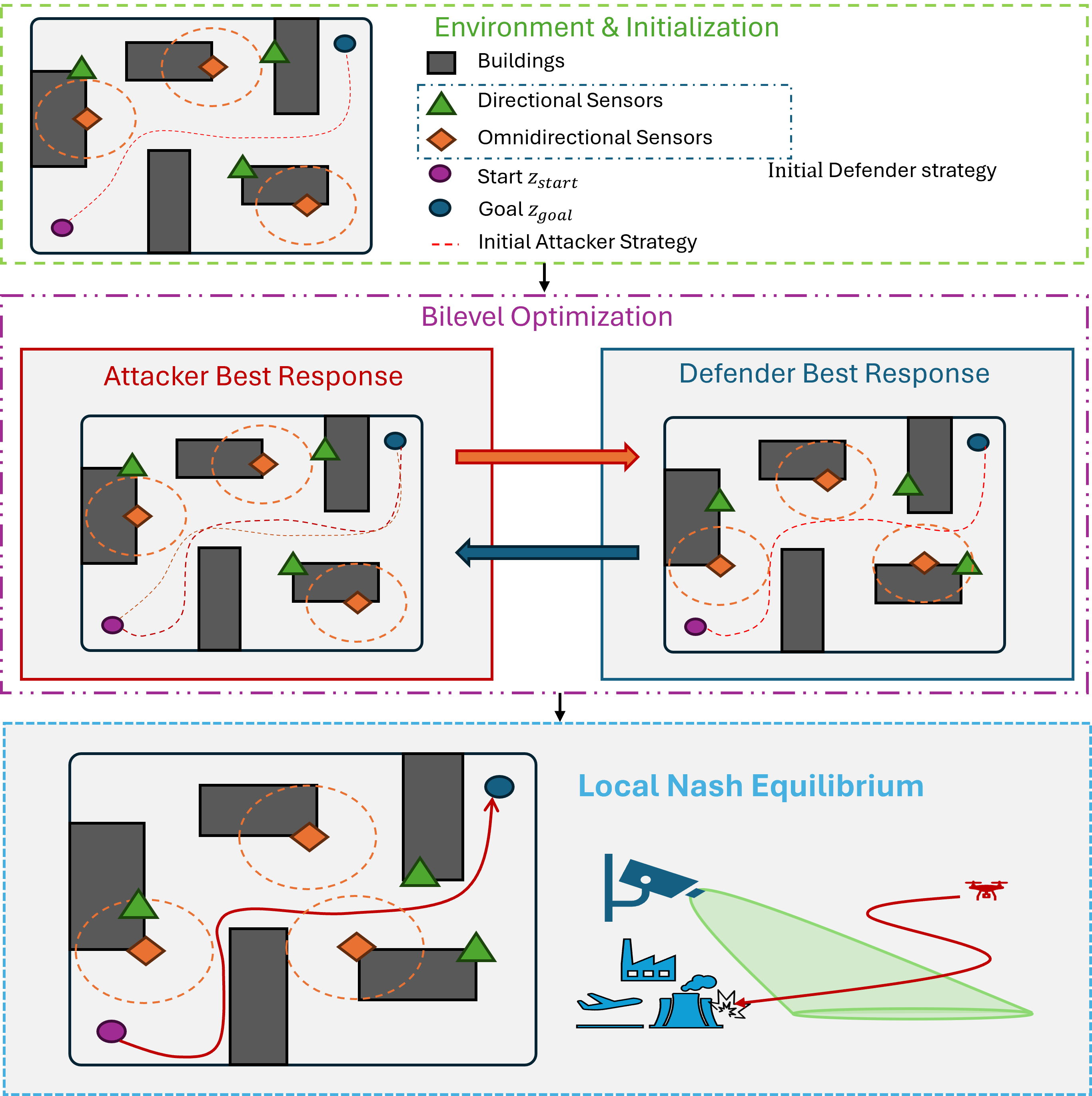}
    \caption{Overview of the proposed bilevel optimization framework. The attacker and defender alternately solve best-response subproblems, minimizing and maximizing detection probability, until convergence to a LNE is achieved.}
    \label{fig:framework_workflow}
\end{figure}

The literature addresses this from three directions.
First, sensor placements \cite{springer_optimal_camera_placement, thesis_optimal_camera, CUAS_03} optimize asset locations to maximize static detection area. However, they fail to consider rotating directional sensors or adversaries exploiting spatiotemporal gaps. Omnidirectional coverage approaches \cite{CUAS_optimization_01, CUAS_optimization_02, CUAS_optimization_03} treat coverage as a convex, time-invariant problem. Ishat-E-Rabban et al. \cite{rabban2021resilient} address adversarial coverage but restrict robots to discrete trajectory sets, making their approach NP-hard, i.e. not solvable in polynomial time. Second, game-theoretic formulations introduce an intelligent adversary. Pirani et al. \cite{CUAS_03} and Zou et al. \cite{Bhattacharya} provide adversarial foundations but restrict placement to discrete graphs or linear action spaces. Fridovich-Keil et al. \cite{fridovich2020efficient} propose iterative linear-quadratic (LQ) approximations for multi-player differential games but omit sensor placement constraints and spatiotemporal detection models. Perimeter defense games \cite{shishika2020cooperative, shishika2018local, Cartee_2019} yield optimal strategies via game-theoretic decomposition, with recent extensions to adaptive positioning \cite{macharet2020adaptive} and multi-vehicle environments \cite{bajaj2024multivehicle}. However, defenders are constrained by binary detection models to fixed configurations and perimeter boundaries. Third, space-time path-planning methods model the attacker's navigation through dynamic sensing obstacles \cite{ICCAS}. Zhang et al. \cite{zhang2021game} extend this to a two-player zero-sum game but restrict both players to discrete grid-based strategy spaces.
Our work combines all three gaps: continuous sensor redeployment along building boundaries, probabilistic directional field of view (FOV) modeling, and gradient-based bilevel optimization over continuous strategy spaces.

We model this the adversarial interaction between the UAS intruder and the sensor network as a differential zero-sum surveillance evasion (SE) game.
We focus on the spatial deployment of sensors as the primary defense variable, holding the surveillance pattern fixed. This isolates the contribution of deployment topology, a necessary first step before addressing the substantially more complex coupled schedule-placement problem.
The attacker's strategy is a collision-free spatiotemporal trajectory from a start point to a goal point that minimizes the probability of detection when the defender has committed to a sensor placement strategy.

We propose an iterative numerical approach to reach a Local Nash Equilibrium (LNE), in which neither player can improve their individual payoff with infinitesimal unilateral deviations.
The acquired LNE solution provides the optimal defense strategy for securing the perimeter against an adversary.

The main contributions of this paper are:
\begin{itemize}
    \item A two-player \emph{zero-sum differential game} formulation that models time-varying, non-convex detection regions of rotating directional sensors over continuous strategy spaces.
    \item A continuous defender redeployment formulation where sensors slide freely along building boundaries, enforced via a log-sum-exp smooth boundary constraint enabling gradient-based optimization.
    \item An iterative bilevel optimization converging to a local minimum satisfying Local Nash equilibrium conditions by alternately optimizing sensor placement and attacker trajectory, with an analytical first-order stationarity condition derived for both players, validated through numerical analysis and experimental demonstration.
\end{itemize}

The remainder of the paper is organized as follows. Section \ref{sec: Problem Formulation} presents the game formulation. Section \ref{sec: SEGame} details the attacker's and defender's best responses, and convergence analysis. The numerical results and conclusions are presented in Sections \ref{sec: Results} and \ref{sec: Conclusion}, respectively.

\section{Problem Formulation}
\label{sec: Problem Formulation}
We formulate the interaction between the attacker and the defender as a two-player zero-sum differential game. Let $\Omega = \mathcal{X} \times \mathcal{T}\subset \mathbb{R}^d$ denote the operational domain, where $\mathcal{X}\subset\mathbb{R}^{d_s}$ is the spatial domain, $\mathcal{T}=[0, T_{\max}]$ is the time horizon.
The attacker aims to navigate from a start point $z_{init}$ to a goal point $z_{goal}$ while minimizing its probability of detection. On the contrary, the defender aims to maximize this probability by selecting optimal sensor placement. We will now present the dynamics and constraints for both the attacker and defender, followed by the zero-sum differential game formulation.

\subsection{Dynamics and Constraints} \label{subsec: dynamics and constraints}
\emph{Attacker Model:} The attacker is a holonomic point-mass agent with state $z(t)\in\Omega$. We define the attacker state as:
\begin{equation}
    \dot{z}(t) = a(t), \quad  z(0) = z_{init}
\end{equation}
where $a(t)\in \mathbb{R}^d$ denotes the velocity control input. The admissible strategy set $\mathcal{A}$ comprises all piecewise continuous velocity functions $a: [0, T_a] \rightarrow \mathbb{R}^d$ satisfying:
\begin{equation}
    ||a(t)||_2 \leq v_{max}, \quad z(T_a) = z_{goal}
\end{equation}
where $v_{\max}$ is the maximum allowable speed and $T_a$ is the mission time horizon within which the attacker must reach $z_{goal}$.

\emph{Defender Model:} The defender operates a heterogeneous network of $M=m_D+m_O$ sensors, consisting of $m_D$ directional and $m_O$ omnidirectional units. Within each type, sensors share identical physical characteristics. Omnidirectional sensors detect uniformly within a range-limited circular region; directional sensors are additionally constrained by a limited angular FOV.

To maximize spatial coverage, each directional sensor $i$ executes a periodic pan motion. This sweeping motion traces a helical path through $\Omega$, generating a time-varying, non-convex coverage region that constitutes the primary evasion challenge for the attacker. Formally, the pan angle $\psi_i(t)$ follows a triangular scan-reversal waveform:
\begin{equation}
    \psi_i(t) = \psi_0 + \frac{2\Delta \psi}{\pi} \arcsin\left(\sin\left(\frac{2\pi t}{T}\right)\right)
    \label{eq: pan_angle}
\end{equation}
where $\psi_0$ is the initial heading, $\Delta \psi$ is the half-sweep angle, and $T$ is the scanning period. The resulting coverage region of sensor $i$ in $\Omega$ is defined as:
\begin{equation}
\mathcal{F}_i = \{z \in \Omega : z \text{ is detectable by sensor } i \text{ at time } t\}
\end{equation}

The defender's strategy $\mathbf{d}\in \mathcal{D}$ is a selection of spatial placement for all $M$ sensors, defined as $\mathbf{d} = \{\mathbf{x}_i, \phi_i(t)\}^M_{i=1}$, where $\mathbf{x_i}\in\Omega$ is the position of sensor $i$. The surveillance pattern $\phi_i(t)$ of each sensor is held fixed throughout the game.
The full feasible strategy space is thus $\mathcal{D}=\{\mathbf{x}_i\}_{i=1}^M$, a continuous set subject to application-specific placement constraints defined in Section \ref{subsec: Defender BR}.

\subsection{Payoff Function and Game Setup} \label{subsec: payoff function and game setup}

The payoff function $J(\mathbf{a}, \mathbf{d})$ represents the cumulative detection probability along the attacker's trajectory.
We model detection as statistically independent across time steps and sensors, i.e., each sensor independently determines whether the attacker can be detected at each time step \cite{fov_ref, tharmarasa2007large}.

We discretize the attacker's trajectory into $N$ time steps, where $k\in\{0, \dots, N-1\}$ denotes the discrete time index and $t_k = k\cdot\Delta t$ is the corresponding continuous time, with $\Delta t = T_f/N$, where $T_f>0$ is the total mission duration.
The cumulative detection probability is:
\begin{equation}
    J(\mathbf{a}, \mathbf{d}) =P_{\text{d}} = 1 -\prod_{k=0}^{N-1}(1 - K(\mathbf{z}_k,S, t_k)) 
    \label{eq: payoff}
\end{equation}
where $S=\{\mathbf{x}_1,\dots,\mathbf{x}_M\}$ is the set of sensor positions, and $K(\mathbf{z}_k, S, t_k) \in [0, 1]$ is the combined instantaneous detection probability at point $\mathbf{z}_k$ and time $t_k$, defined formally in Section \ref{subsec: Attacker BR}.

The game value is:
\begin{equation}
    V = \min_{\mathbf{a} \in \mathcal{A}} \max_{\mathbf{d} \in \mathcal{D}} J(\mathbf{a}, \mathbf{d})
\end{equation}
The strategy space $\mathcal{A}$ and $\mathcal{D}$ are continuous and infinitely-dimensional. Furthermore, $J$ is neither convex in $\mathbf{a}$ nor concave in $\mathbf{d}$, so a global saddle point need not exist \cite{Huang, minimax_optimization}. Classical finite-action solvers are therefore inapplicable to this infinite game \cite{kamra}.

We therefore seek a LNE, a tractable solution concept for continuous non-convex games \cite{minimax_optimization}. A strategy pair $(\mathbf{a}^*, \mathbf{d}^*)$ constitutes an LNE \cite{Local_Nash_Eq} if there exist neighborhoods $\mathcal{N}_a\subset\mathcal{A}$ containing $\mathbf{a}^*$ and $\mathcal{N}_d\subset\mathcal{D}$ containing $\mathbf{d}^*$ such that:

\begin{equation}
J(\mathbf{a}^*, \mathbf{d}) \leq J(\mathbf{a}^*, \mathbf{d}^*) \leq J(\mathbf{a}, \mathbf{d}^*)
\label{eq: LNE}
\end{equation}

for all $\mathbf{a} \in \mathcal{N}_a$ and $\mathbf{d} \in \mathcal{N}_d$. This means neither player can improve their individual payoff through unilateral deviations from their current strategy.

In theory, the optimal solution is characterized by the Hamilton-Jacobi-Isaacs (HJI) equation, which is intractable in our setting for three reasons. First, the time-varying, non-convex coverage regions of rotating directional sensors produce discontinuities in the value function \cite{bansal2017hamilton, chen2018hamilton}. Second, the joint continuous strategy spaces of both players yield a high-dimensional PDE without closed-form solutions \cite{bacsar1998dynamic}. Third, the nonlinear sigmoid visibility model $\Psi_i$ and the product structure of $K(\mathbf{z}, S, t)$ render both subproblems non-convex. Convex relaxations under limited assumptions, such as linearizing the FOV model, would undermine the physical characteristics of the detection model. We therefore adopt a bilevel optimization approach that separates the minimax problem into two alternating subproblems, detailed in Section \ref{sec: SEGame}.

\section{Surveillance Evasion Game} \label{sec: SEGame}
We present the bilevel optimization framework for computing the LNE of the SE game. Following conventions in security games \cite{Tambe, Kiekintveld, Bustamante}, the defender commits to a sensor placement before the attacker responds. The framework alternates between two best-response subproblems: the attacker minimizes detection probability given the current defender placement, and the defender maximizes it given the current attacker trajectory. Section \ref{subsec: attacker and defender strategy} formalizes the iterative scheme and convergence conditions. Section \ref{subsec: Attacker BR} and Section \ref{subsec: Defender BR} detail the attacker and defender optimization subproblems, respectively.

\subsection{Attacker and Defender Strategy} \label{subsec: attacker and defender strategy}
We separate the minimax problem in Equation \ref{eq: LNE} into two alternating nonlinear programs (NLPs): an attacker subproblem that minimizes cumulative detection probability over trajectory waypoints, and a defender subproblem that maximizes it over sensor placements. The iterative scheme converges to a first-order stationary point consistent with the LNE condition rather than a global saddle point, which is computationally impractical in continuous strategy spaces \cite{Facchinei}. Throughout this paper, $(\cdot)^*$ denotes a first-order stationary point of the respective player's subproblem, not a globally optimal solution.

\emph{Attacker strategy: }Given a fixed, fully known defender placement $\mathbf{d}$, the game reduces to a single-sided optimal control problem. Since the detection model $K(\mathbf{z}, S, t)$ is non-convex due to the nonlinear sigmoid visibility function $\Psi_i$, a convex simplification is precluded. We therefore adopt a two-step approach: first, STP-RRT* \cite{ICCAS} generates a feasible collision-free initial trajectory through the space-time domain $\Omega$, navigating around dynamic sensor coverage regions; second, a NLP improves this trajectory to minimize $P_d$, solved via IPOPT \cite{wachter2006ipopt}. The NLP initialization from STP-RRT* is critical. Without a feasible starting point, gradient-based solvers risk falling into poor local minima in the non-convex detection landscape.  The attacker seeks a locally optimal trajectory $\mathbf{a}^*$ minimizing cumulative detection probability:
\begin{equation}
    \mathbf{a}^* = \arg\min_{\mathbf{a} \in \mathcal{A}}( 1-\prod_{i=0}^{N-1}(1-K(\mathbf{z}_i,S, t_i))
\end{equation}
subject to the dynamics and boundary constraints in Section \ref{subsec: dynamics and constraints}.

\emph{Defender strategy: } Given the attacker's locally optimal strategy $\mathbf{a}^*$, the defender solves a continuous sensor placement problem to maximize detection probability along the attacker's path:
\begin{equation}
    \mathbf{d}^* = \arg\max_{\mathbf{d} \in \mathcal{D}}( 1-\prod_{i=0}^{N-1}(1-K(\mathbf{z}_i,S, t_i))
    \label{eq: dstar}
\end{equation}
subject to the application-specific position constraints, e.g., building edges for urban or airport defense scenarios.
In contrast to prior game-theoretic approaches that restrict the defender to discrete placement graphs or fixed configurations \cite{Cartee_2019, CUAS_03}, our formulation allows sensors to slide continuously along physical boundaries. This enables adaptive coverage of spatiotemporal gaps as the attacker's strategy evolves, with boundary placement enforced via a smooth differentiable constraint detailed in Section \ref{subsec: Defender BR}.

\emph{Iterative bilevel scheme: } The bilevel scheme proceeds by alternating between the two best-response subproblems. One complete round consists of the attacker's best response followed by the defender's best response. At iteration $k$, strategies are updated as:
\begin{equation}
    \mathbf{a}^{k+1} \leftarrow \arg\min_{\mathbf{a}\in\mathcal{A}} J(\mathbf{a}, \mathbf{d}^k)
\end{equation}
\begin{equation}
    \mathbf{d}^{k+1} \leftarrow \arg\max_{\mathbf{d}\in\mathcal{D}} J(\mathbf{a}^k, \mathbf{d})
\end{equation}
where $\mathbf{a}^k\in\mathcal{A}$ and $\mathbf{d}^k\in\mathcal{D}$ denote the attacker and defender strategies at iteration $k$, respectively. Each subproblem is solved to a first-order stationary point, and the alternating updates drive both players toward mutual best-response. Since the game is zero-sum with a single payoff $J(\mathbf{a}, \mathbf{d})$, we track $J_A^k = J(\mathbf{a}^k, \mathbf{d}^{k-1})$ and $J_D^k = J(\mathbf{a}^{k-1}, \mathbf{d}^{k})$ separately to monitor each player's improvement against the opponent's most recent strategy.

\emph{Convergence and stationarity: } The iterative scheme terminates when both players reach a first-order stationary point. As shown in Section \ref{subsec: Attacker BR}, $J(\mathbf{a}, \mathbf{d})$ is continuously differentiable with respect to both $\mathbf{a}$ and $\mathbf{d}$ due to the smooth sigmoid detectability model $K(\mathbf{z}, S, t)$, allowing the first-order stationarity conditions for the LNE $(\mathbf{a}^*, \mathbf{d}^*)$ to be stated analytically:
\begin{equation}
    \nabla_{\mathbf{a}} J(\mathbf{a}^*, \mathbf{d}^*) = \mathbf{0}, \quad \nabla_{\mathbf{d}} J(\mathbf{a}^*, \mathbf{d}^*) = \mathbf{0}
    \label{eq: stationarity}
\end{equation}

In practice, exact satisfaction of Equation \ref{eq: stationarity} is verified numerically by monitoring the gradient residuals of both subproblems. The iteration terminates when:
\begin{equation}
    \max\left(||\nabla_{\mathbf{a}} J||, ||\nabla_{\mathbf{d}} J||\right) < \delta
    \label{eq: stopping}
\end{equation}
where $\delta > 0$ is a prescribed tolerance. Furthermore, the alternating scheme may exhibit limit cycle behavior, characterized by non-decaying oscillation in $\Delta J$ without satisfying Equation \ref{eq: stopping}. This occurs when multiple local minima of comparable cost exist in proximity. If the maximum number of iterations $K_{\max}$ is reached without convergence, the defender reinitializes from a new random placement $\mathbf{d}^0$ and the optimization restarts.
\begin{algorithm}
\small
\caption{Bilevel SE Game Optimization}
\label{alg:bilevel}
\begin{algorithmic}[1]
\Require Initial attacker trajectory $\mathbf{a}^0$ via STP-RRT*, 
         maximum iterations $K_{\max}$, tolerance $\delta$, 
         maximum reinitializations $R_{\max}$
\Ensure LNE strategy pair $(\mathbf{a}^*, \mathbf{d}^*)$
\State $r \leftarrow 0$
\Repeat
    \State Sample random defender placement $\mathbf{d}^0$ along building edges
    \State $k \leftarrow 0$
    \Repeat
        \State $\mathbf{a}^{k+1} \leftarrow \arg\min_{\mathbf{a} \in \mathcal{A}} J(\mathbf{a}, \mathbf{d}^k)$ 
        \State $\mathbf{d}^{k+1} \leftarrow \arg\max_{\mathbf{d} \in \mathcal{D}} J(\mathbf{a}^k, \mathbf{d})$ 
        \State $k \leftarrow k + 1$
    \Until{$\max(||\nabla_{\mathbf{a}} J||, ||\nabla_{\mathbf{d}} J||) < \delta$ \textbf{or} $k \geq K_{\max}$}
    \If{$\max(||\nabla_{\mathbf{a}} J||, ||\nabla_{\mathbf{d}} J||) < \delta$}
        \State \textbf{return} $(\mathbf{a}^*, \mathbf{d}^*) \leftarrow (\mathbf{a}^k, \mathbf{d}^k)$
    \EndIf
    \State $r \leftarrow r + 1$ 
\Until{$r \geq R_{\max}$}
\State \textbf{return} best $(\mathbf{a}^k, \mathbf{d}^k)$ found across all reinitializations
\end{algorithmic}
\end{algorithm}
Algorithm \ref{alg:bilevel} summarizes the full bilevel optimization procedure. The inner loop alternates between the attacker and defender best-response subproblems until the stopping criterion in Equation \ref{eq: stopping} is satisfied. If the maximum number of iterations $K_{\max}$ is reached without convergence, the defender reinitializes from a new random placement and the inner loop restarts. This process repeats until convergence is achieved or the maximum number of reinitializations $R_{\max}$ is reached.

\subsection{Attacker Best Response} \label{subsec: Attacker BR}
We formulate the attacker's best response as the following NLP:
\begin{equation}
\label{eq: optimization_problem}
\begin{split}
    \min_{\mathbf{Z}, S} \quad & J_A(\mathbf{Z}, S) = -\sum_{k=0}^{N-1} \ln \left(1 - K(\mathbf{z}_k, S, t_k)\right)\\
    \text{s.t.} \quad & \mathbf{z} \in \mathcal{X} , \quad \forall k \in \{0, \dots, N-1\} \\
    & \mathbf{z}_{k+1} = \mathbf{z}_k + \mathbf{v}_k \Delta t, \quad \forall k \in \{0, \dots, N-1\} \\
    & \|\mathbf{v}_k\|_2 \leq v_{\text{max}}, \quad \forall k \in \{0, \dots, N-1\} \\
    & \mathbf{z}_0 = \mathbf{z}_{\text{start}}, \quad \mathbf{z}_N = \mathbf{z}_{\text{goal}} \\
    & T_f \leq T_a
\end{split}
\end{equation}
where $\mathbf{z}_k\in\mathcal{X}\subset\mathbb{R}^{d_s}$ is the attacker's spatial position at step $k$, $\mathbf{Z}=\{\mathbf{z}_0,\dots,\mathbf{z}_N\}$ is the discretized trajectory, $\Delta t=T_f/N$, and $\mathbf{z}_k=(\mathbf{z}_{k+1}-\mathbf{z}_k)/\Delta t$ is the mean velocity.
Although $\Omega=\mathcal{X}\times\mathcal{T}$ is defined over space-time, the NLP decision variables are restricted to $\mathcal{X}$, with time handled explicitly via $T_f$ and $\Delta t$.

To ensure continuous differentiability for gradient-based NLP solvers, we adopt a smooth pointwise detectability model. For sensor $i$ at $\mathbf{x}_i$ with time-varying heading $\phi_i(t)$, the detectability is:
\begin{equation}
\label{eq: point-wise_detectability}
K_i(\mathbf{z},  \mathbf{x}_i, t) = D_i(\mathbf{z}) \cdot \Psi_i(\mathbf{z}, t)
\end{equation}
where $D_i(\mathbf{z})$ is a distance decay function and $\Psi_i(\mathbf{z}, t)$ is a directional visibility function, defined as:
\begin{align}
D_i(\mathbf{z}) &= \frac{1}{\alpha |\mathbf{x}_i - \mathbf{z}|^2 + 1} + \sigma \\
\label{eq: sigmoid_vis}\Psi_i(\mathbf{z}, t) &= \frac{1}{1 + \exp\left( -c \left( \cos \gamma - \cos(\theta/2) \right) \right) }
\end{align}
where $\alpha$ is a distance scaling constant, $\sigma$ is a minimal detection bias, $\theta$, $c$ are the FOV angle and sigmoid sharpness parameter, and $\cos\gamma(t)=\hat{\mathbf{r}}_j\cdot\mathbf{h}_j(t)$ is the angular deviation between the FOV centerline and the vector from $\mathbf{x}_i$ to $\mathbf{z}$, where $\hat{\mathbf{r}}_j = (\mathbf{z}_k-\mathbf{x}_j)/|\mathbf{z}_k-\mathbf{x}_j|$ is the unit vector from sensor $j$ to the attacker. The time dependence of $K_i$ arises entirely through the sensor heading $\mathbf{h}_i(t)=[\cos\phi_i(t), \sin\phi_i(t)]^T$, which follows the triangular scan pattern.

The combined detectability across all $M$ sensors, consistent with Equation \ref{eq: payoff} is:
\begin{equation}
    K(\mathbf{z}, S, t) = 1-\prod_{j=1}^M (1-K_j(\mathbf{z}, \mathbf{x}_j, t))
    \label{eq: total detectability}
\end{equation}
For omnidirectional sensors, $\Psi_i$ is treated as constant.

For large $N$, the product $\Pi_{k=0}^{N-1}(1-K(\mathbf{z}_k, S, t_k))$ approaches zero faster than double-precision floating point can represent, causing numerical underflow. We therefore equivalently minimize $-\sum_{k=0}^{N-1} \ln (1-K(\mathbf{z}_k, S, t_k))$, from which the detection probability is recovered as: 
\begin{equation}
\label{eq: p_detected}
    P_{d} = 1 - \exp\left(\sum_{k = 0}^{N-1}\ln(1 -K(\mathbf{z}_k,S, t_k))\right)
\end{equation}

\subsection{Defender Best Response} \label{subsec: Defender BR}

Unlike the attacker, whose search space is the open operational domain $\Omega$, the defender's candidate positions are constrained by the physical environment: sensors must remain mounted on building surfaces.
Unlike prior discrete~\cite{Cartee_2019} or fixed-deployment~\cite{CUAS_03} approaches, we formulate a continuous redeployment problem where each sensor slides freely along the edges of convex polygons.

We model each building $B$ as a convex polygon with intersection of $M_e$ half-spaces. Each sensor $j\in\{1,\dots,M\}$ is assigned to a building, and its position $\mathbf{x}_j\in\mathbb{R}^2$ must satisfy:
\begin{equation}
    \mathbf{A}_j\mathbf{x}_j\leq \mathbf{b}_j
\end{equation}
where $A_j\in\mathbb{R}^{M_e\times2}$ and $b_j\in\mathbb{R}^{M_e}$ represent the linear inequalities defining the building's geometry.

To allow sensors to slide continuously along all edges of the assigned building, we require each sensor to lie exactly on the polygon boundary. A naive hard boundary constraint is non-differentiable at the vertices, which makes it incompatible with gradient-based NLP solvers. We therefore enforce boundary placement via a log-sum-exp smooth approximation of the signed distance to the nearest edge, as illustrated in Figure \ref{fig:lse_boundary}:
\begin{figure}[h!]
    \centering
    \includegraphics[width=0.95\linewidth]{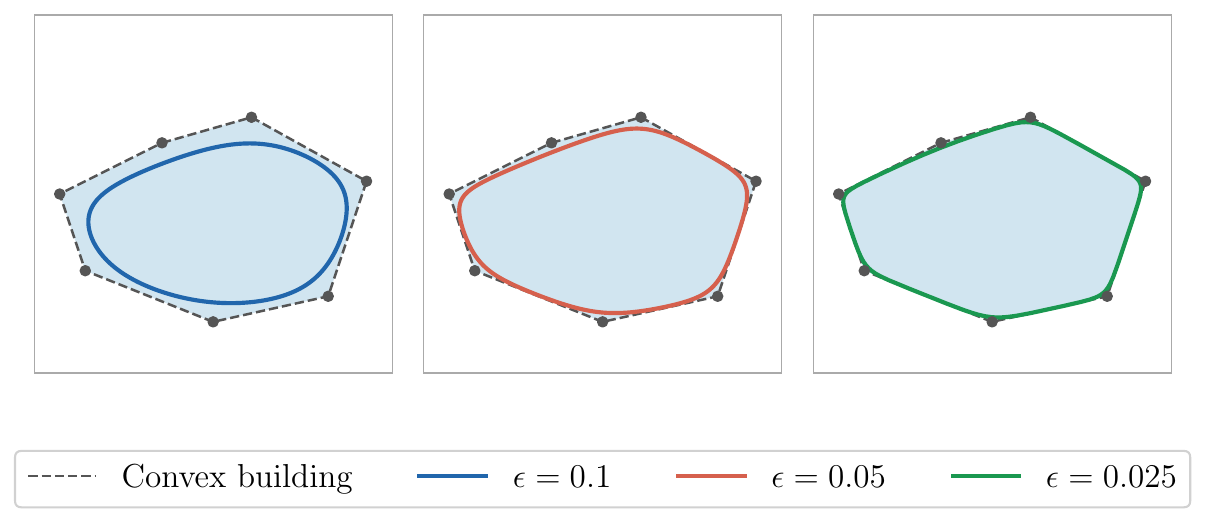}
    \caption{Log-sum-exp smooth approximation $\mathcal{R}(\mathbf{x}_j)=0$ of a convex polygon building boundary at three values of $\epsilon$. As $\epsilon\rightarrow0$, the approximation recovers the exact boundary, enabling gradient-based sensor redeployment along building edges.}
    \label{fig:lse_boundary}
\end{figure}
\begin{equation}
    \mathcal{R}(\mathbf{x}_j) = -\epsilon \ln \sum_{k=1}^{M_e} \exp\left({\frac{\mathbf{A}^T_{k,j}\mathbf{x}_j-\mathbf{b}_{k,j}}{\epsilon}}\right) = 0
    \label{eq: binary_residual}
\end{equation}
where $\epsilon > 0$ is a smoothing parameter, controlling the tradeoff between smoothness and constraint accuracy. The log-sum-exp formulation remains continuously differentiable across the full boundary, recovering the exact signed distance as $\epsilon\rightarrow 0$.


Given a fixed attacker trajectory $\mathbf{Z}=\{\mathbf{z}_0, \dots,\mathbf{z}_N\}$, the defender solves:
\begin{equation}
    \label{eq:defender_optimization}
    \begin{split}
        \max_{\mathbf{x}_1,\dots,\mathbf{x}_M} \quad & J_D(\mathbf{Z},S)= \sum_{k=0}^{N-1} - \ln (1-K(\mathbf{z}_k, \mathbf{S}, t_k)) \\
        \text{s.t.} \quad & \mathbf{A}_j \mathbf{x}_j \leq \mathbf{b}_j,\quad \forall j \in \{1,\dots,M\} \\
        & \mathcal{R}(\mathbf{x}_j) = 0, \quad \forall j \in {1,\dots,M}
    \end{split}
\end{equation}
where $J_D$ is the cumulative log-odds of detection over the attacker's full trajectory. By maximizing $J_D$, the defender sensor placements that most effectively take advantage of vulnerabilities in the attacker's current trajectory, closing the spatiotemporal gaps the attacker attempted to exploit.

\subsection{Convergence to local minima}
\label{subsec:convergence_LNE}
Due to the smooth sigmoid $\Psi_j$ and differentiable distance decay $D_j$, $K(\mathbf{z}, \mathbf{x}_i, t)$ is continuously differentiable with respect to both $\mathbf{z}$ and $\mathbf{x}_j$. Thus, the first-order stationarity conditions for both subproblems can be stated analytically. The attacker's stationarity condition is:
\begin{equation}
    \nabla_{\mathbf{z}_k}J_A(\mathbf{Z}^*, S^*) = \sum_{k=0}^{N-1} \sum_{j=1}^{M} \frac{\Psi_j\nabla_{\mathbf{z}_k} D_j + D_j\nabla_{\mathbf{z}_k}\Psi_j}{1-K_j(\mathbf{z}_k^*, \mathbf{x}_j^*, t_k)}
\end{equation}
where component gradients are defined as:
\begin{align}
    \nabla_{\mathbf{z}_k} D_j &= \frac{-2\alpha(\mathbf{z}_k - \mathbf{x}_j)}{(\alpha|\mathbf{x}_j - \mathbf{z}_k|^2 + 1)^2} \\
    \nabla_{\mathbf{z}_k} \Psi_j &= \frac{c\,\Psi_j(1-\Psi_j)}{|\mathbf{z}_k - \mathbf{x}_j|}
    \left[\mathbf{h}_j(t) - (\mathbf{h}_j(t)\cdot\hat{\mathbf{r}}_j)\hat{\mathbf{r}}_j\right]
\end{align}

Since $\nabla_{\mathbf{x}_j}\Psi_j = -\nabla_{\mathbf{z}_k}\Psi_j$ and $\nabla_{\mathbf{x}_j}D_j = -\nabla_{\mathbf{z}_k} D_j$, the defender gradient satisfies $\nabla_{\mathbf{x}_j}K_j = -\nabla_{\mathbf{z}_k}K_j$. 
The full defender KKT stationarity condition, with active boundary constraints, is:
\begin{multline}
    -\sum_{k=0}^{N-1} \frac{\Psi_j \nabla_{\mathbf{z}_k} D_j + 
    D_j \nabla_{\mathbf{z}_k} \Psi_j}{1 - K_j(\mathbf{z}_k^*, \mathbf{x}_j^*, t_k)} 
    + \lambda_j \mathbf{A}_j^T + \mu_j \nabla_{\mathbf{x}_j}\mathcal{R}(\mathbf{x}_j^*) = 0, \\
    \forall j \in \{1,\dots,M\}
\end{multline}
where $\lambda_i\geq0$ are KKT multipliers for the half-space constraints $\mathbf{A}_j\mathbf{x}_j\leq\mathbf{b}_j$, $\mu_j$ are multipliers for the boundary residual $\mathcal{R}(\mathbf{x}_j^*)=0$, and:
\begin{align}
\nabla_{\mathbf{x}_j}\mathcal{R}(\mathbf{x}_j) &= 
-\frac{\displaystyle\sum_{k=1}^{M_e} \mathbf{A}_{k,j}
\exp\left(\frac{\mathbf{A}_{k,j}^T\mathbf{x}_j - \mathbf{b}_{k,j}}{\epsilon}\right)}
{\displaystyle\sum_{k=1}^{M_e}
\exp\left(\frac{\mathbf{A}_{k,j}^T\mathbf{x}_j - \mathbf{b}_{k,j}}{\epsilon}\right)}
\end{align}

The pair $(\mathbf{Z}^*, S^*)$ constitutes LNE when both stationarity conditions are satisfied simultaneously. Since finding a global NE in a continuous strategy space is computationally impractical \cite{Facchinei}, we verify stationarity numerically using the stopping criterion:
\begin{equation}
    \max(||\nabla_{\mathbf{Z}}J_A(\mathbf{Z}^*, S^*)||, ||\nabla_S J_D(\mathbf{Z}^*, S^*) ||) < \delta
\end{equation}

\begin{figure}
    \centering
    \includegraphics[width=0.95\linewidth]{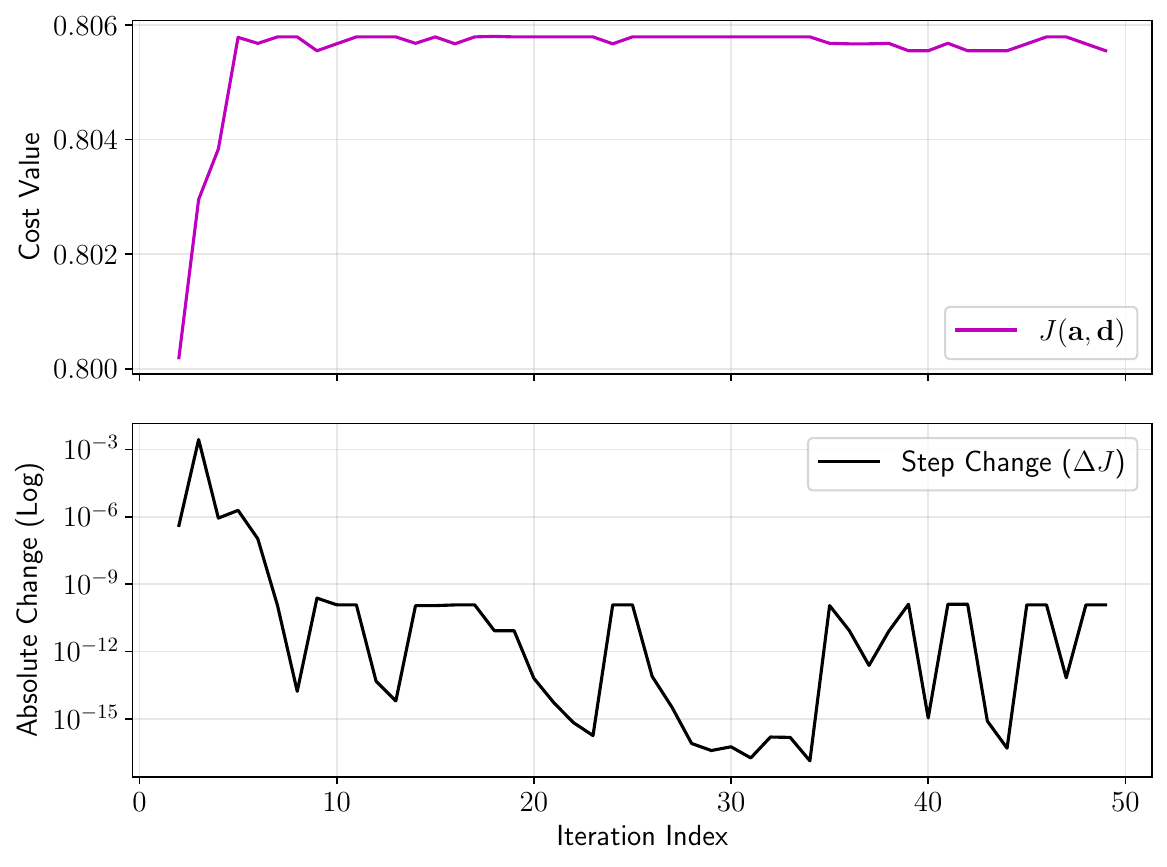}
    
    \caption{A representative convergence trace of the alternating bilevel optimization. (Top) Joint payoff $J(\mathbf{a}^k, \mathbf{d}^k)$ evaluated at each iteration's strategy pair. (Bottom) Change of numerical residual $\Delta J$.}
    \label{fig: convergence}
\end{figure}

Figure \ref{fig: convergence} illustrates a representative convergence of the alternating optimization. The joint payoff $J(\mathbf{a}^k, \mathbf{d}^k)$, evaluated at each iteration's strategy pair, stabilizes into a plateau. This indicates mutual best response. The numerical residual decays to near $10^{-6}$, confirming first-order stationarity. Convergence behavior across all $500$ trials is summarized in Table \ref{tab:success_rate}.

\section{Results and Analysis}
\label{sec: Results}
We evaluate the proposed framework on a CUAS mission scenario with a heterogeneous sensor network deployed around convex building obstacles. All simulations are implemented in Python using CasADi \cite{andersson2018casadi} with IPOPT \cite{wachter2006ipopt} as the NLP solver.

\subsection{Simulation Setup}
\label{subsec: sim_setup}
The operational domain $\mathcal{X}$ is a $100\times100$ grid with convex polygon buildings randomly placed as obstacles. The sensor network consists of $M=10$ sensors: 5 directional and 5 omnidirectional, each mounted on building surfaces. Directional sensors follow the triangular scan pattern defined in \eqref{eq: pan_angle} with parameters drawn from the ranges in Table \ref{tab:sim_parameters}. The attacker navigates from a fixed start point $\mathbf{z}_{start}$ to a fixed goal point $\mathbf{z}_{goal}$, with maximum speed $v_{\max}$ as defined in Table \ref{tab:sim_parameters}. The bilevel optimization is initialized with the attacker trajectory from STP-RRT* and a randomized defender placement along building edges, and terminates when the stopping criterion in Equation \ref{eq: stopping} falls below $\delta = 10^{-5}$.

\begin{table}[htbp]
    \centering
    \begin{tabular}{l c c}
        \toprule
        \textbf{Parameter} & \textbf{Case Study} & \textbf{Monte Carlo} \\
        \midrule
        $\psi_{\text{init}}$ & $[0, 2\pi)$ & $[0, 2\pi)$ \\
        $v_{\text{pan}}$ (deg/sec) & $[-10, 10]$ & $[-10, 10]$ \\
        Map size ($x, y$) & $100 \times 100$ & $150 \times 150$ \\
        Range $R$ (units) & $30$ & $[10, 15]$ \\
        Sensors $M$ & $5$ (each type) & $5$--$10$ (each type) \\
        FOV $\theta$ (deg) & $30$ & $[10, 30]$ \\
        \bottomrule
    \end{tabular}
    \caption{Simulation parameters for the case study and Monte Carlo evaluation.}
    \label{tab:sim_parameters}
\end{table}

\subsection{SE Game Case Studies}
\label{subsec: case_studies}
\begin{figure}[h!]
    \centering
    \includegraphics[width=0.95\linewidth]{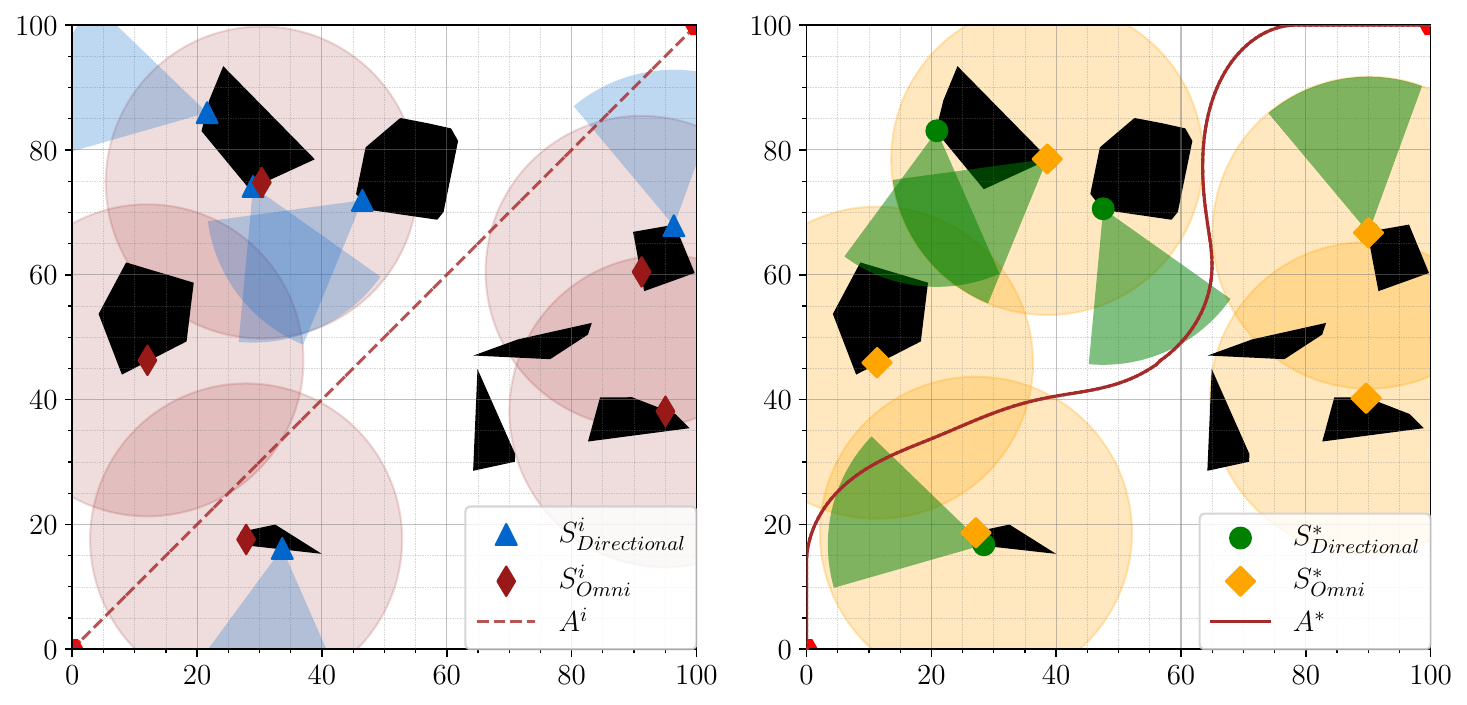}
    \caption{Case study of the SE game example with 5 directional and omnidirectional sensors deployed in the operation area. (Left) Initial strategies for attacker and defender. (Right) Optimized strategies for attacker and defender.}
    \label{fig:case1}
\end{figure}
Figure \ref{fig:case1} illustrates the LNE strategies for a representative environment. The left panel shows the initial randomized defender placement $\mathbf{d}^0$ and the STP-RRT* attacker trajectory $\mathbf{a}^0$. The right panel shows the converged LNE strategies $(\mathbf{a}^*, \mathbf{d}^*)$ after bilevel optimization.

At initialization, the attacker's trajectory passes through regions of low sensor coverage, exploiting gaps between sensor FOVs. The defender's randomized placement leaves significant spatiotemporal gaps along this path. After optimization, the attacker's trajectory $\mathbf{a}^*$ reroutes through the central corridor, increasing lateral distance from sensors to minimize $P_d$. In response, the defender's sensors translate along building edges to close these corridors, creating overlapping FOV coverage at choke points. The converged defender placement raises the attacker's minimum achievable detection probability across all available paths, confirming mutual best-response at the LNE.

\subsection{Performance Evaluation}
\label{subsec: Performance}
We conducted a $500$-trial Monte Carlo simulation, with uniform randomly selected parameters shown in Table \ref{tab:sim_parameters}, and a fixed start and goal position.
On average, each trial required $96.7$ seconds on a standard workstation, with the attacker's STP-RRT* initialization accounting for most of the computation.
The defender's NLP was solved via IPOPT in CasADi \cite{andersson2018casadi}, converging in 12 iterations on average per bilevel round.



\begin{table}[h]
    \centering
    \footnotesize
    \begin{tabular}{l c c c}
    \toprule
    & \textbf{Initial} $J^0 \pm 95\%$ \textbf{CI} 
    & \textbf{LNE} $J^* \pm 95\%$ \textbf{CI} 
    & $\Delta$ \textbf{(Mean} $\pm$ \textbf{95\% CI)} \\
    \midrule
    $J_A$ & $0.402 \pm 0.019$ & $0.361 \pm 0.017$ & $-0.040 \pm 0.006$ \\
    $J_D$ & $0.103 \pm 0.011$ & $0.419 \pm 0.020$ & $+0.316 \pm 0.016$ \\
    \bottomrule
    \end{tabular}
    \caption{Monte Carlo performance summary comparing $J_A$ and $J_D$ at initialization and LNE.}
    \label{tab:success_rate}
\end{table}

Table \ref{tab:success_rate} reports $J_A^k$ and $J_D^k$ at initialization and convergence, reflecting each player's payoff against the opponent's most recent strategy. The defender achieves a 4$\times$ improved detection probability at LNE ($\Delta J_D = +0.316\pm0.016$), confirming convergence toward a local Nash Equilibrium. The randomized initial placement $\mathbf{d}^0$ leaves significant room for optimization, which the bilevel optimization consistently exploits across all $500$ trials.

\begin{figure}[htbp]
\centering
\begin{minipage}{\columnwidth}
    \centering
    \includegraphics[width=0.95\columnwidth]{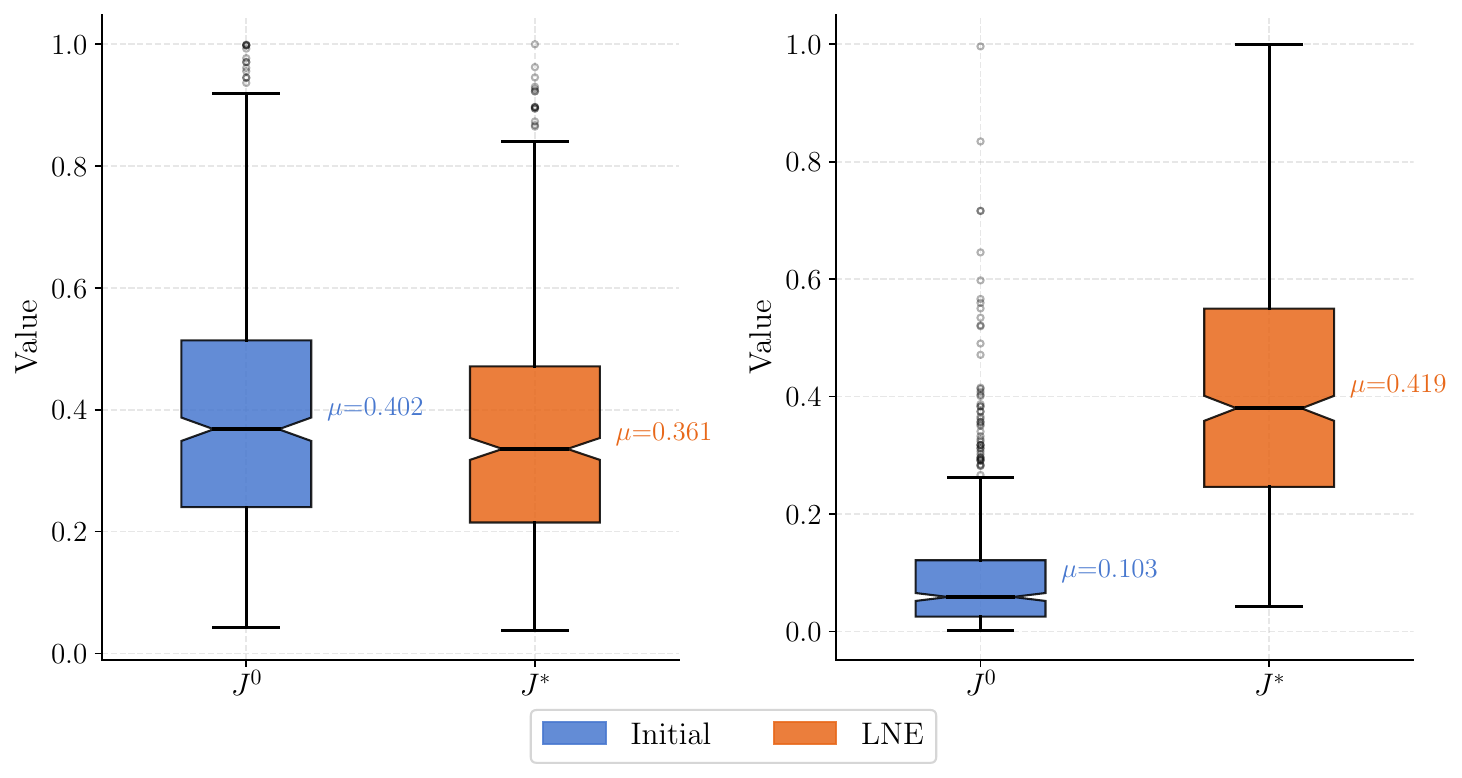}
    \label{fig:boxplot}
\end{minipage} 

\caption{500-Trial Monte Carlo simulation results comparing distributions of $J_A$ and $J_D$ under randomized initial strategies $J^0$ and LNE strategies $J^*$.}
\label{fig:monte_carlo_result}
\end{figure}

Figure \ref{fig:monte_carlo_result} illustrates the distribution of $J_A$ and $J_D$ across all trials. The wide interquartile range of $J_D^0$ concentrated near zero reflects the inefficiency of random sensor placement, while $J_D^*$ shows a significant upward shift, confirming that the bilevel optimization consistently drives the defender toward higher detection probability, even with an inefficient initial condition.

\begin{figure}[htbp]
    \centering
    \includegraphics[width=0.95\columnwidth]{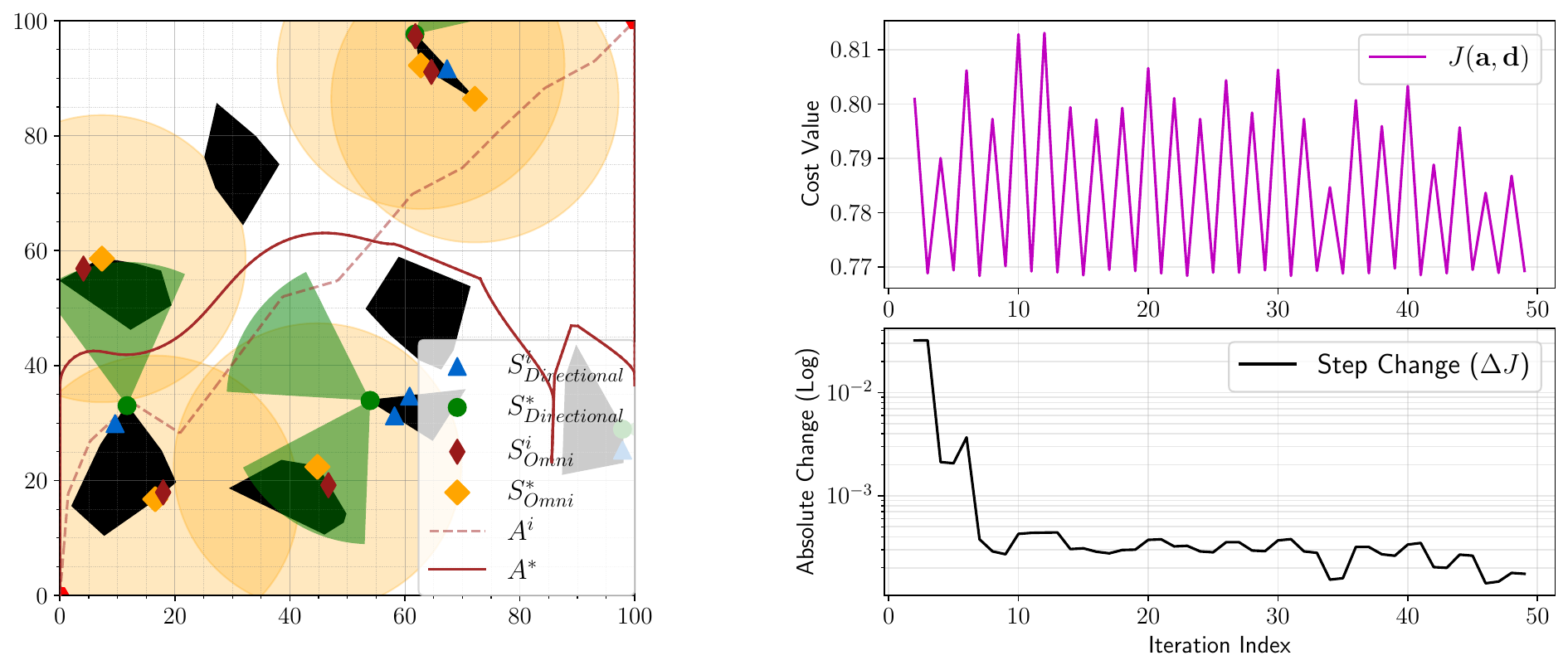}
    \caption{Limit cycle behavior in the bilevel optimization. (Left) The attacker and the defender's selected strategies are shown in the operation map. (Right) The oscillation of strategy cost, showing failure of convergence on LNE}
    \label{fig:limit_cycle}
\end{figure}

Limit cycle behavior, characterized by non-decaying oscillation in $\Delta J$ without satisfying the stationarity condition, was observed in $1$ out of $500$ trials ($2 \%$) as illustrated in Fig \ref{fig:limit_cycle}. In this case, reinitialization from a new random defender placement successfully recovered convergence, confirming the robustness of the proposed heuristic, even in diverse environments.

\subsection{Experimental Demonstration}
\begin{figure}[hbt!]
    \centering
    \includegraphics[width=0.95\columnwidth]{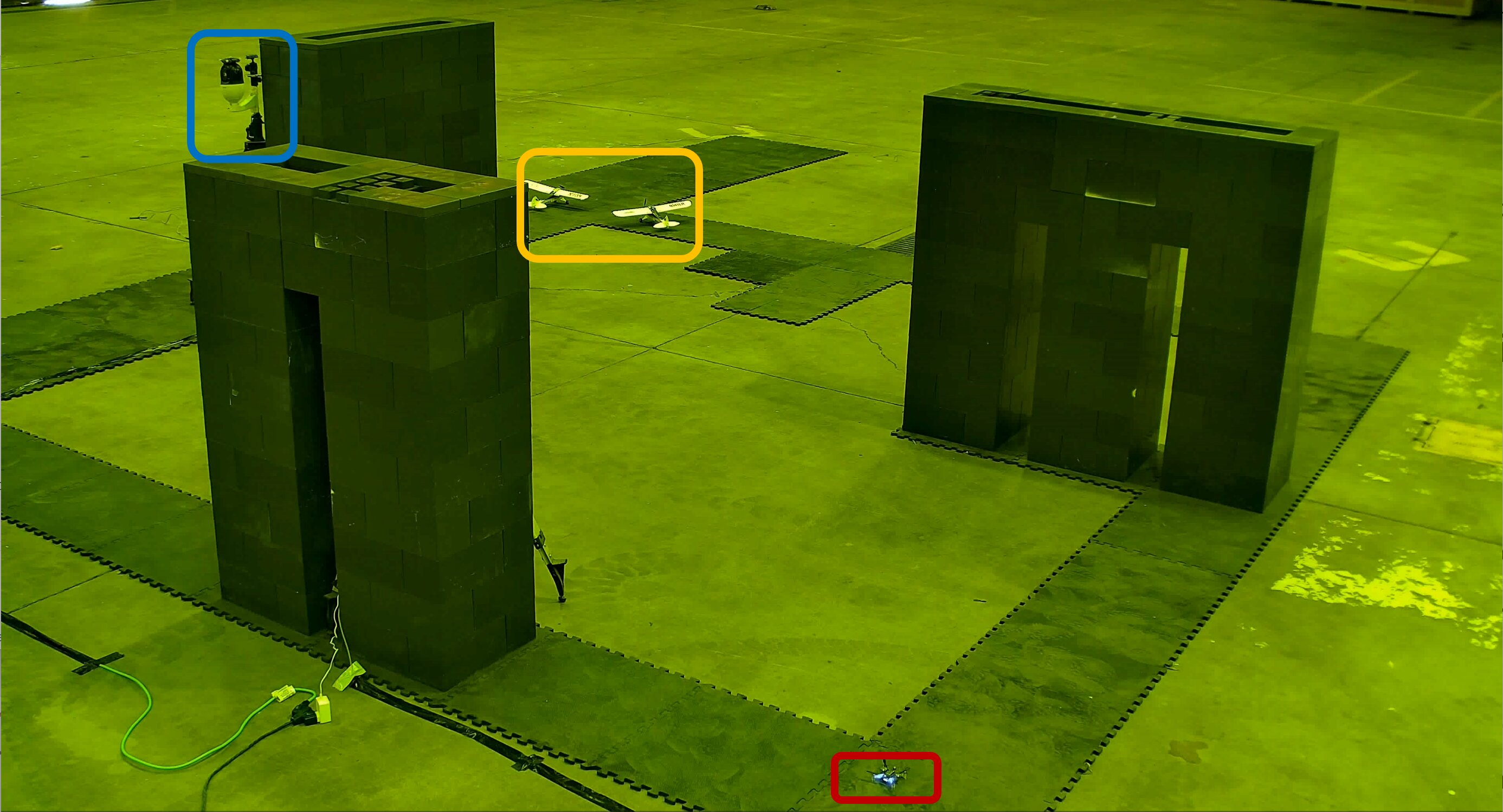}
    \caption{Experimental setup with the urban canyon mockup simulating an airfield environment. The Reolink RLC-823A PTZ camera (defender, blue), aircrafts (target, orange), and Crazyflie 2.1 (adversarial UAS, red) are labeled.}
    \label{fig:exp_setup}
\end{figure}

We demonstrated the framework at a UAS test facility, simulating an airport defense scenario. A scaled-down mockup of an airport strip with an urban canyon environment was constructed. A Crazyflie 2.1 \cite{giernacki2017crazyflie} served as the attacker, and three Reolink RLC-823A x16 PTZ cameras served as the defender. Four scenarios were evaluated: 1) initial heuristic $(\mathbf{a}^0, \mathbf{d}^0)$, 2) defender best-response $(\mathbf{a}^0, \mathbf{d}^*)$, 3) attacker best-response $(\mathbf{a}^*, \mathbf{d}^0)$, and 4) LNE strategy $(\mathbf{a}^*, \mathbf{d}^*)$. Fig.~\ref{fig:exp_setup} shows the physical test environment constructed for the demonstration.

\begin{figure}[hbt!]

\begin{subfigure}{.475\linewidth}
  \includegraphics[width=\linewidth, trim={9.8cm 1.8cm 5.8cm 4cm}, clip]{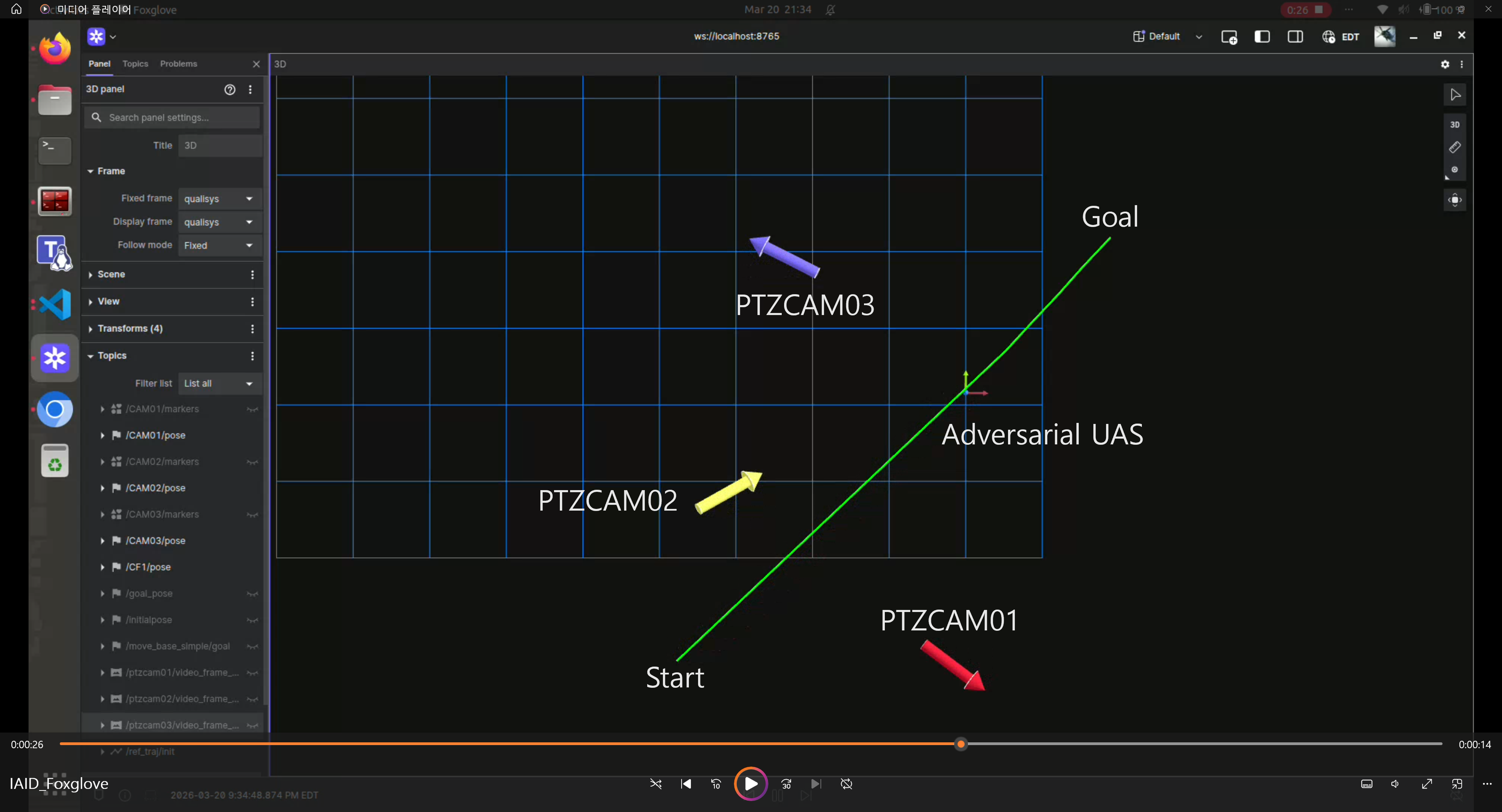}
  \caption{$(\mathbf{a}^0, \mathbf{d}^0)$}
  \label{fig:demo_init}
\end{subfigure}\hfill 
~ 
\begin{subfigure}{.475\linewidth}
  \includegraphics[width=\linewidth, trim={9.8cm 1.8cm 5.8cm 4cm}, clip]{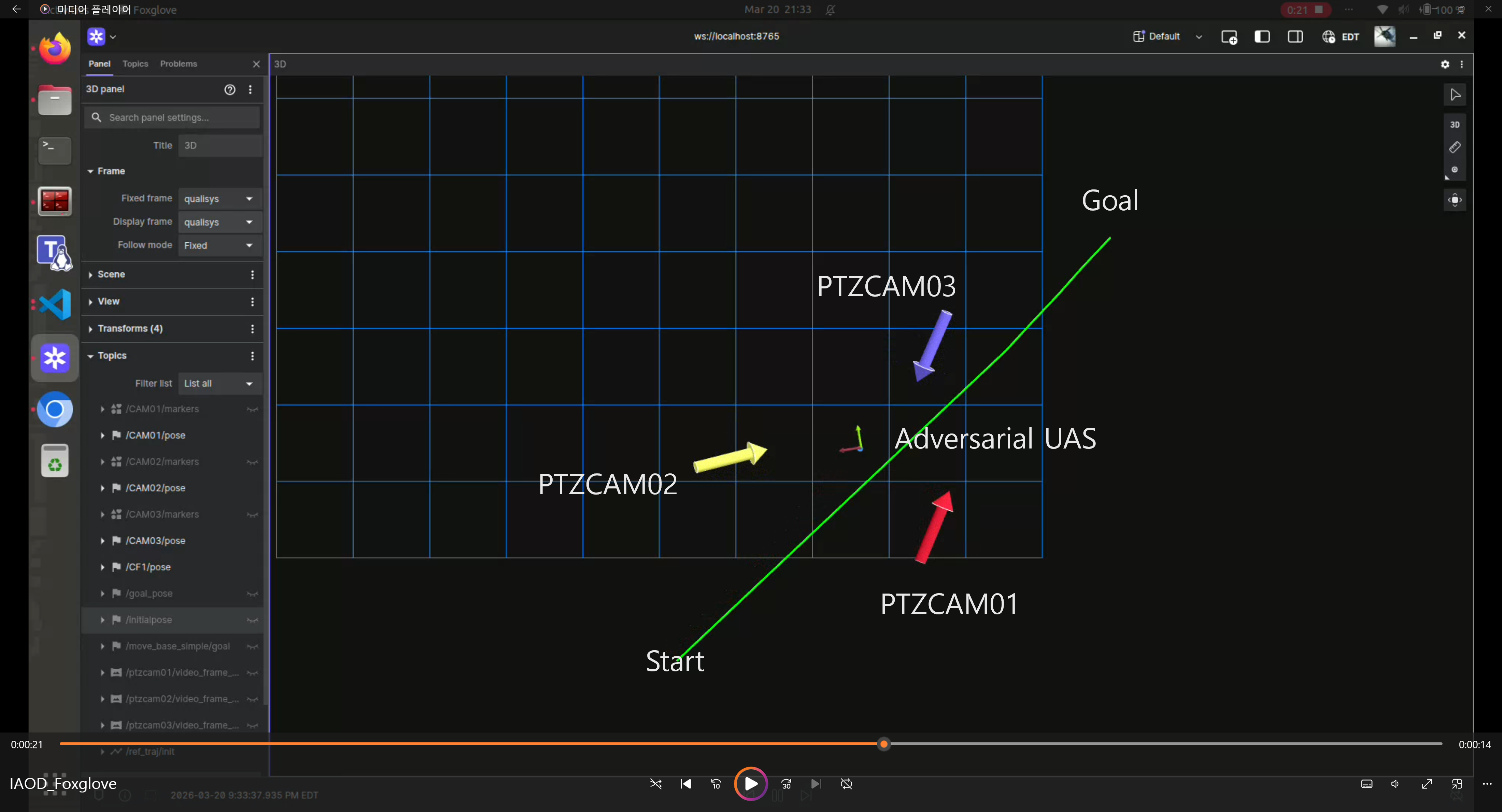}
  \caption{$(\mathbf{a}^0, \mathbf{d}^*)$}
  \label{fig:demo_DBR}
\end{subfigure}
\vspace{-0.3cm}
\medskip 
\begin{subfigure}{.475\linewidth}
  \includegraphics[width=\linewidth, trim={9.8cm 1.8cm 5.8cm 4cm}, clip]{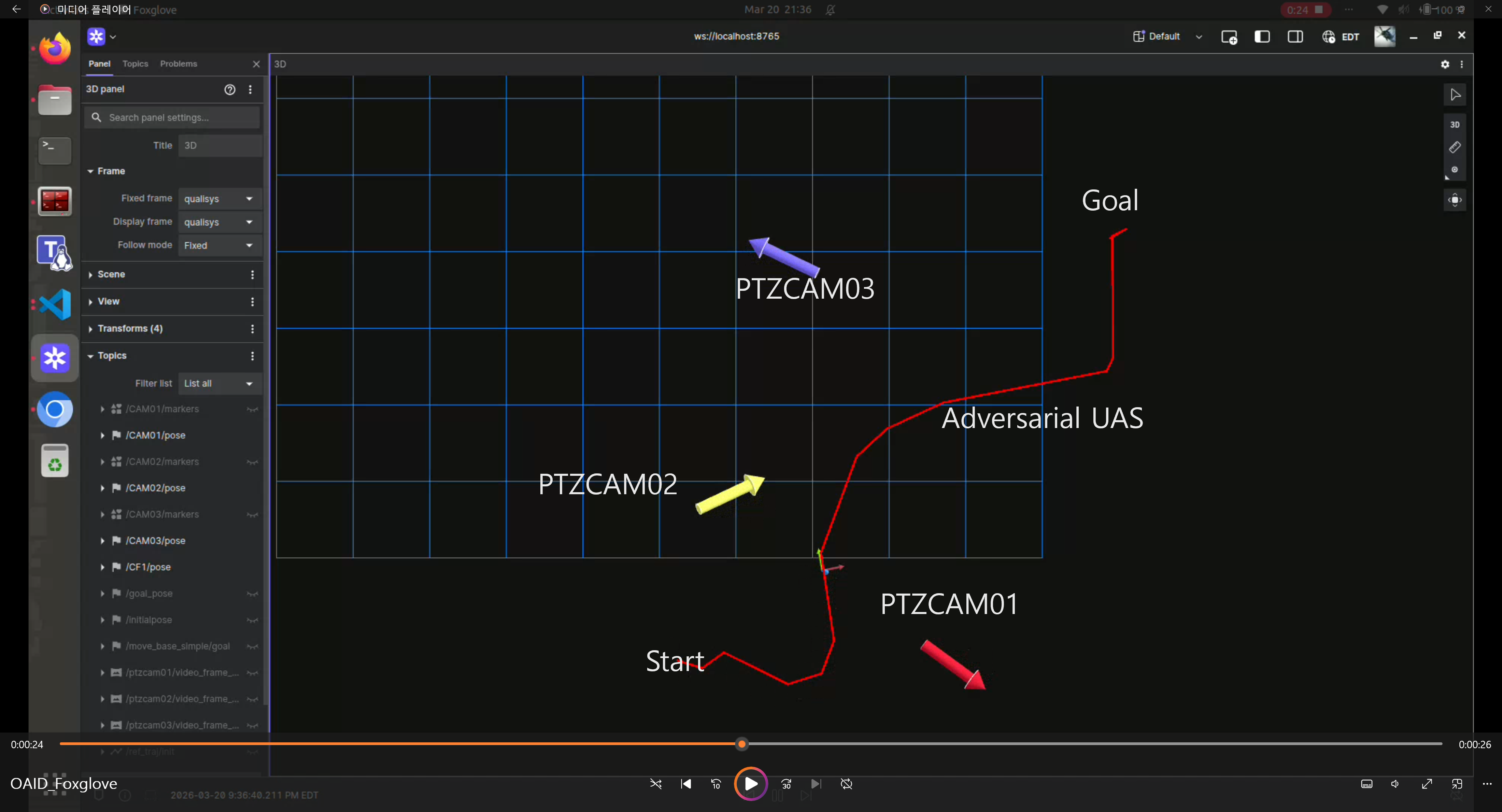}
  \caption{$(\mathbf{a}^*, \mathbf{d}^0)$}
  \label{fig:demo_ABR}
\end{subfigure}\hfill 
\begin{subfigure}{.475\linewidth}
  \includegraphics[width=\linewidth, trim={9.8cm 1.8cm 5.8cm 4cm}, clip]{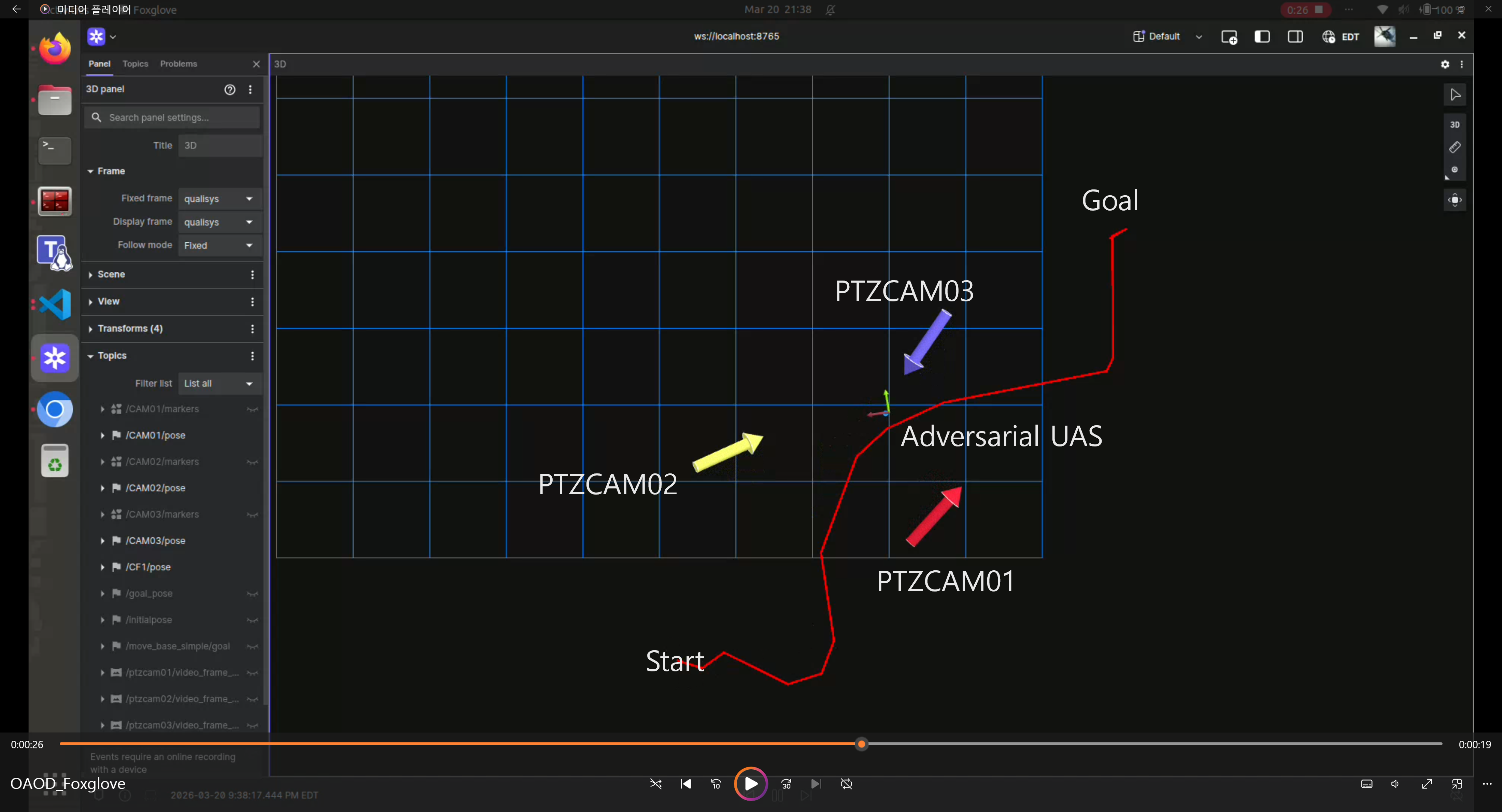}
  \caption{$(\mathbf{a}^*, \mathbf{d}^*)$}
  \label{fig:demo_LNE}
\end{subfigure}
\caption{Experimental demonstration of the surveillance evasion framework. PTZCAM01 (red), PTZCAM02 (yellow), and PTZCAM03 (purple) arrows indicate sensor headings. Heuristic and optimized attacker trajectories are shown in green and red, respectively.}
\label{fig:demo}
\end{figure}

\begin{table}[h]
    \centering
    \begin{tabular}{lccccc}
    \toprule
    \textbf{ } & $(\mathbf{a}^0, \mathbf{d}^0)$ & $(\mathbf{a}^0, \mathbf{d}^*)$ & $(\mathbf{a}^*, \mathbf{d}^0)$ & $(\mathbf{a}^*, \mathbf{d}^*)$ \\
    \midrule
    $J$ & 0.6847& 0.9587 & 0.6711  & 0.9451 \\
    \bottomrule
    \end{tabular}
    \caption{Experimental Demonstration Results}
    \label{tab:exp_result}
\end{table}

Figure \ref{fig:demo} shows the four scenarios. Sensors PTZCAM01, PTZCAM02, and PTZCAM03 are shown as red, yellow, and purple arrows, respectively, and the attacker's trajectory is overlaid as a colored path. In $(\mathbf{a}^0, \mathbf{d}^0)$, the attacker follows a heuristic straight-line path while sensors remain in their initial placement, leaving significant coverage gaps along the diagonal corridor. In $(\mathbf{a}^*, \mathbf{d}^0)$, the optimized attacker reroutes to take advantage of these gaps, reducing $J$ against the same defender. In $(\mathbf{a}^0, \mathbf{d}^*)$, the defender repositions sensors to close the attacker's original path, maximizing $J$ against the same attacker. At LNE $(\mathbf{a}^*, \mathbf{d}^*)$, both players commit to their optimal strategies, confirming mutual best-response. Table \ref{tab:exp_result} reports the computed detection cost $J$ across all four scenarios, validating that the bilevel optimization produces consistent strategy improvements in a real-world setting.



\section{CONCLUSIONS}
\label{sec: Conclusion}

We introduce a game-theoretic framework for surveillance evasion in counter-UAS missions via bilevel optimization over continuous strategy spaces. The framework models the interaction between an adversarial UAS and a heterogeneous sensor network as a two-player zero-sum differential game that converges to a Local Nash Equilibrium, where neither player can unilaterally improve their payoff. A 500-trial Monte Carlo simulation demonstrates a $4\times$ improvement in defender detection probability over randomized sensor placement, the natural baseline given the absence of directly comparable prior methods, with a $96.8\%$ convergence rate. Future work includes joint sensor scheduling, nonholonomic attacker dynamics, and 3D extensions.


\section*{ACKNOWLEDGMENT}
The authors used Grammarly and Anthropic's Claude AU to improve the grammar and syntax of several paragraphs in this manuscript, in accordance with IEEE RAS guidelines for generative AI usage.


\bibliographystyle{ieeetr}
\bibliography{IEEEabrv,refs}

\addtolength{\textheight}{-12cm}   
\end{document}